\documentclass{article}
\usepackage{kr}

\usepackage{times}
\usepackage{soul}
\usepackage{url}
\usepackage[hidelinks]{hyperref}
\usepackage[utf8]{inputenc}
\usepackage[small]{caption}
\usepackage{subcaption}
\usepackage{graphicx}
\usepackage{tabularx}
\usepackage{amsmath}
\usepackage{amsthm}
\usepackage{amsfonts}
\usepackage{amssymb}
\usepackage{booktabs}
\usepackage{algorithm}
\usepackage{algpseudocode}
\usepackage{enumitem}
\usepackage{multirow}
\usepackage{diagbox}
\usepackage{listings}
\usepackage{xcolor}

\lstdefinelanguage{Prolog}{
    morekeywords={:-, ?, is, not, fail, true, listing}, 
    sensitive=true,
    morecomment=[l]{\%}, 
    morestring=[b]",
}

\usepackage[section]{placeins}
\usepackage{array}

\newcommand{\gpt}{GPT-4o}
\newcommand{\gt}{Game theory}

\graphicspath{{./pics/}}

\title{Autoformalization of Game Descriptions using Large Language Models}

\author{
Agnieszka Mensfelt$^1$\and
Kostas Stathis$^1$\and
Vince Trencsenyi$^1$\and \\
\affiliations
$^1$Department of Computer Science, Royal Holloway, University of London\\
\emails
\{agnieszka.mensfelt, kostas.stathis, vince.trencsenyi\}@rhul.ac.uk 
}

\begin{document}

\maketitle

\begin{abstract}
Game theory is a powerful framework for reasoning about strategic interactions, with applications in domains ranging from day-to-day life to international politics. However, applying formal reasoning tools in such contexts is challenging, as these scenarios are often expressed in natural language. To address this, we introduce a framework for the autoformalization of game-theoretic scenarios, which translates natural language descriptions into formal logic representations suitable for formal solvers. Our approach utilizes one-shot prompting and a solver that provides feedback on syntactic correctness to allow LLMs to refine the code. We evaluate the framework using \gpt{} and a dataset of natural language problem descriptions, achieving 98\% syntactic correctness and 88\% semantic correctness. These results show the potential of LLMs to bridge the gap between real-life strategic interactions and formal reasoning.
\end{abstract}

\section{Introduction}

Game theory~\cite{vonNeumann1944} is a mathematical framework that facilitates the analysis of competitive and cooperative interactions among rational decision-makers. It can be applied to a wide range of scenarios, from family decision-making to business strategies to nuclear conflict. However, these scenarios are typically expressed in natural language, making it challenging to apply formal reasoning tools directly. Moreover, the inherent ambiguities and complexities of natural language complicate the task of automatically converting problem descriptions into formal representations suitable for solvers.

Recent advancements in Large Language Models (LLMs) offer a solution to this challenge. LLMs exhibit remarkable abilities in handling the nuances of natural language in the context of translation. They have proved to be successful in converting natural language into formal representations, including mathematical~\cite{wu2022autoformalization,he2023solving,jiang2022draft} and logical expressions~\cite{cosler2023nl2spec,pan2023logic,feng2023language,yang2023coupling,chen2023nl2tl}, which is known as autoformalization~\cite{wu2022autoformalization}. This paper explores the novel application of LLMs in the context of game theory, specifically focusing on translating natural language descriptions of game-theoretic scenarios into formal representations that enable formal reasoning in real-world situations.

To the best of our knowledge, this work is the first to leverage LLMs for autoformalization within the domain of game theory. The results pave the way for more easily accessible applications of game theory, bridging the gap between natural language and formal reasoning in strategic interactive scenarios. The main contributions of this paper are as follows:

\begin{itemize}
    \item \textbf{Dataset creation}.  We developed a dataset consisting of 105 natural language descriptions of scenarios that can be modelled using game theory. This dataset includes both standard and non-standard descriptions, with and without numerical payoffs, to vary the translation difficulty level.
    \item \textbf{Autoformalization framework development}. We propose a novel framework for translating natural language descriptions into formal representations, enabling formal reasoning in real-world game-theoretic scenarios.
    \item \textbf{Evaluation}. We evaluate the performance of \gpt{} within our framework by assessing its ability to translate natural language descriptions of strategic interactions into formal representations. Our evaluation includes both zero-shot and one-shot prompting and examines the model's capacity to generate formal specifications for games that differ structurally from provided examples.
\end{itemize}

The remainder of this paper is organized as follows. Section~\ref{sec:background} provides a background overview, including LLMs, game theory, and general game playing. Section~\ref{sec:methods} details the formal solver, the proposed framework, experimental setup, and the dataset of game descriptions. Section~\ref{sec:results} presents the evaluation results on the dataset. Section~\ref{sec:related-work} reviews related work. Finally, Section~\ref{sec:conclusions} provides the conclusions and outlines future work.

\section{Background}
\label{sec:background}

\paragraph{Large Language Models}
Multi-layered LSTMs are sequential models that perform well on complex sequential tasks such as language translation \cite{NIPS2014_a14ac55a}. Following the emergence of transformers and the success of pre-trained models \cite{Qiu2020}, such architectures had become the go-to solution for state-of-the-art \textit{Large Language Models (LLMs)}. LLM performance depends on billions of parameters and huge training data sets ~\cite{zhao2023survey}. While LLMs generalize well over a wide variety of tasks, fine-tuning can be used to optimize their performance for specific tasks ~\cite{ziegler2020finetuning}.

\subsubsection{\gpt{}}
OpenAI's \gpt{}~\cite{gpt4o2024} is a current state-of-the-art LLM. Its multi-modal architecture supports text, image and audio inputs and outputs. \gpt{} supersedes GPT-4 ~\cite{achiam2023gpt} and is estimated to have ~175B parameters. It has been shown to perform exceptionally well in reasoning tasks \cite{wang2024rupbenchbenchmarkingreasoningperturbations}.

\paragraph{Game Theory}

    \begin{table}
    \renewcommand{\arraystretch}{1.15}
        \centering
        \caption{A general game in matrix form, with the row and column players payoffs recorded as a tuple: $(\pi_{row},\pi_{column})$}
        \begin{tabular}{r|cc}
            \textbf{P1/P2} & \textbf{Cooperate} & \textbf{Defect} \\
            \hline
            \textbf{Cooperate} & (R, R) & (S, T) \\
            \textbf{Defect} & (T, S) & (P, P) \\
        \end{tabular}
        \label{subtab:PD}
    \end{table}

\gt{} provides a mathematical framework to reason about strategic interactions and decision-making analytically \cite{Osborne2004}. Following the formalization by \cite{OsborneRubinstein1994}, we define a game theoretic encounter -- a game -- by its players, the available strategies and payoffs.

\paragraph{Players}

\gt{} assumes a simple definition of rationality to hold for all of its actors -- the players -- and that each actor is aware that all others are also rational in the same sense. Such players are expected to act intelligently and voluntarily, with a full understanding of the game and to be acting to optimize their own utility score \cite{myerson1984introduction}.

\paragraph{Strategies}
A player's strategy maps the state of the game to an available action, also considering the accumulated history \cite{OsborneRubinstein1994}.

\paragraph{Payoff}
A payoff is a numerical value associated with a specific outcome of strategic interaction; it is the reward each player receives as a result of a combination of their actions \cite{Gibbons1992}.

\paragraph{Game}

Based on the outlines of \cite{OsborneRubinstein1994} and \cite{Rasmusen2004introtogametheory}, our games are formally defined in the following way:
\begin{itemize}
    \item $N$: the set of $n$ players;
    \item $A_i$: $\forall i \in N$, a non-empty set of actions available to the player;
    \item $u_i$: a utility function mapping $A_i \rightarrow \mathbb{R}$, $\forall i \in N$, producing $\pi_i$.
\end{itemize}

In the context of our experiments, we refer to non-cooperative games in normal form \cite{Rasmusen2004introtogametheory}. A general example is shown in Table ~\ref{subtab:PD}. Our five games are characterised by such $2\times2$ payoff matrices, having two players and two actions -- \textit{Cooperate} and \textit{Defect}. Taking the row player's perspective, these actions lead to four possible outcomes: 
\begin{itemize}
    \item \textbf{T}: the \textit{Temptation to defect}, resulting from the opponent choosing D against the player's C;
    \item \textbf{R}: the \textit{Reward for mutual cooperation};
    \item \textbf{P}: the \textit{Punishment for mutual defection};
    \item \textbf{S}: the \textit{Sucker's Payoff}, resulting from the player cooperating while the opponent chooses to defect.
\end{itemize}

\noindent The relation of these four outcomes can be used to classify strategic interaction into different games \cite{Rasmusen2004introtogametheory}.

\paragraph{Prisoner's Dilemma}
The Prisoner's Dilemma is a symmetric game defined by a payoff structure that values the temptation to defect over mutual cooperation: $T>R>P>S$. An example payoff matrix is (-1,-1) for mutual cooperation, (0,-10) for the temptation to defect, (-10,0) for the sucker's payoff, and (5,5) for mutual defection. The negative payoffs represent years spent in prison.

\paragraph{Hawk-Dove}
The Hawk-Dove game describes a situation where \textit{T} is associated with the highest reward, but the \textit{punishment for mutual defection} yields the lowest utility. The definitive relation $T>R>S>P$ is demonstrated by the following payoff assignments: \textbf{R}:(0,0), \textbf{S}:(-1,1), \textbf{T}:(1,-1), \textbf{P}:(-10,-10).

\paragraph{Matching Pennies}
Matching Pennies is a zero-sum game. In our generalised version of the game, if both players choose the same action, the row player gets higher payoff while the opponent gets lower payoff. If the player choices are in disagreement, the yields are the reverse resulting in the asymmetric outcome relations: $R,P>T,S$ and $T,S > R,P$.

\paragraph{Stag Hunt}
The Stag Hunt involves a payoff structure which values mutual cooperation over the temptation to defect. For instance, mutual cooperation gives both players a payoff of $3$, the temptation to defect yields $2$ while the sucker gets $0$, and mutual defection will result in a reward of $1$ utility for both -- resulting in the outcome relation $R>T>P>S$.

\paragraph{Battle of the Sexes}
In this asymmetric coordination game, both players have a preferred action, while disagreement is the least preferred outcome for both \cite{Osborne2004}. For instance, mutual cooperation yields $(2,1)$, mutual defection gives $(1,2)$, and distinct player choices result in $0$ for both. From the row player's perspective, the situation is characterised by the relation $R>P>T,S$, and $P>R>T,S$ for the column player.

\paragraph{General Game Playing}

The idea behind general game playing~\cite{gdl} is about how to construct intelligent systems that can process the rules of arbitrary new games and learn to play these games without human intervention. Key to this idea is the notion of a Game Description Language (GDL), proposed as a formal and machine-processable language for describing the rules of arbitrary games~\cite{gdl-orig}. Initially, GDL focused on information games only, but its extended version GDL-II covers both n-player games with incomplete information~\cite{gdl-ii}) and games in extensive normal form~\cite{gdl-univ}. GDL-II is based on the standard syntax and semantics of logic programming and characterised by special keywords about the role of players, what holds in the initial, intermediate, and final states of the game, and other useful predicates about legal moves, and the utility players once a goal state is achieved. 

A challenge with GDL systems is learning without human guidance. To address this challenge players must be able to reason about the possible actions of others, essentially evaluating hypothetical game situations before taking action. Action languages like the classical Situation Calculus~\cite{sc} have been developed for precisely this purpose. Formal schemes and inference methods are readily available for Situation Calculus~\cite{sc-games1,sc-games2}, while their deployment in general game-playing presupposes a translation from GDL into existing, suitably expressive formalisms. One such scheme~\cite{Schiffel_Thielscher_2011} shows how to fully embed GDL-II into a version of the Situation Calculus based on knowledge fluents~\cite{sc-kbf}. Inspired by these works, we present a general game solver in section~\ref{subsec:solver}, that reasons with GDL specifications, and show how the way the solver is represented can guide an LLM to generate the game-specific parts of different games after having seen example.

\section{Methods}
\label{sec:methods}

Our methods rely on a logic programming solver that provides a game game-independent formulation of an initial state, a set of legal moves, the effects of these moves on the state, and a description of the final states of a game. This formulation can generate or test all possible evolutions of a specific game, given we provide it with the domain-dependent part describing the specifics of that game. If we then feed the solver a logic programming description of a specific game and a textual description of that game to an LLM, we can then prompt the LLM to generate the formulations of different games from textual descriptions. We then devise an algorithm that allows us to do just that, viz., automatically formalise logic programs of new games systematically from textual descriptions of these games.

\subsection{Solver}
\label{subsec:solver}

Our solver consists of a game-independent part specifying the rules of any game in extensive form, a game-dependent part expressing the rules of a specific game, and a set of auxiliary predicates used on top of these representations to support processing a game. We represent all parts in Prolog: variables are denoted by uppercase letters, predicates and function symbols by lowercase letters. The symbol \texttt{:-} is read as \texttt{if}, and \texttt{\textbackslash +} is read as \texttt{not} (negation by failure). An underscore \texttt{'\_'} is used to denote a variable whose value is unused within a specific definition. In this setting, the state of a game is represented as a situation understood by histories of moves starting in an initial situation denoted by a constant (e.g. {\tt s0}). The special binary function {\tt do(M, S)} represents the situation resulting from the execution of move {\tt M} in situation {\tt S}, as in the Situation Calculus.

\subsubsection{Game-independent part}

We specify legal transitions of an extended form game from an initial situation {\tt S} to a final situation {\tt F} as:
\begin{lstlisting}
game(F,F):- 
    final(F).
game(S,F):- 
    \+ final(S), 
    legal(M,S), 
    game(do(M,S),F).
\end{lstlisting}
The final situation {\tt F} is returned when reached. In a non-final situation {\tt S}, the game accepts a legal move {\tt M}, and the game continues in the next {\tt do(M,S)} situation, until the final situation {\tt F} is reached. To reason about what holds in a situation, we use Situation Calculus:

\begin{lstlisting}
holds(F, S):- 
    initially(F, S).
holds(F, do(M, S)):- 
    effect(F, M, S).
holds(F, do(M, S)):- 
    holds(F, S), 
    \+ abnormal(F, M, S).
\end{lstlisting}

A fluent {\tt F} holds in the initial situation, a new fluent {\tt F} is initiated by the effects of a move {\tt M} executed in a situation {\tt S}, and a fluent {\tt F} persists after a move is made, provided it is not abnormal; abnormal fluents are terminated (do not persist). Rules of the form:
\begin{lstlisting}
finally(F, S):- Conditions.
\end{lstlisting}
return derived fluents {\tt F} describing the result of the game, when the {\tt Conditions} hold  in the final situation {\tt S}. 

\subsubsection{Game-dependent part}

For a specific game, we define game-dependent predicates for the initial state {\tt initial/1}, the legal moves {\tt legal/2}, what holds in the initial game situation via {\tt initially/2}, the effects of a move on a situation via {\tt effect/3}, what stops persisting in a situation after the execution of a move via {\tt abnormal/3}, the final situation {\tt final/1}, and the result of the game via {\tt finally/2} definitions. To exemplify a game definition, we show how to describe a PD game with initial situation {\tt s0}, defined as:
\begin{lstlisting}
initial(s0).
\end{lstlisting}
What holds in {\tt s0} we specify as:
\begin{lstlisting}
initially(player(p1), s0).
initially(player(p2), s0).
initially(role(p1,row), s0).
initially(role(p2,col), s0).
initially(control(p1), s0).
initially(control(p2), s0).
\end{lstlisting}
Player names are represented by unique identifiers ({\tt p1} and {\tt p2}), their roles ({\tt p1} is the {\tt row} player, while {\tt p2} is the {\tt col}umn player), and then the fact that initially any of them can play next (we use a {\tt control/1} fluent to indicate that, as in GDL). What holds in the initial situation changes as a result of the move being made in the game. We represent moves as terms of the form {\tt choice(P, M)}, where {\tt P} is a player, and {\tt M} is a move. As it is possible for any player in a Prisoner's Dilemma game to choose defect ({\tt 'D'}) or cooperate ({\tt 'C'}), we write this as:
\begin{lstlisting}
possible(choice(P,'D'), S):-
    holds(player(P), S).
possible(choice(P,'C'), S):-
    holds(player(P), S).
\end{lstlisting}
It is then legal for a player to choose a possible move if they have the control to execute it in the current situation:
\begin{lstlisting}
legal(choice(P,M), S):-  
    possible(choice(P,M), S),
    holds(control(P), S).
\end{lstlisting}
When a legal move {\tt M} is made by a player {\tt P}, the effect that this move is actually made is recorded in the next situation:
\begin{lstlisting}
effect(did(P, M), choice(P, M), S).
\end{lstlisting}
Once a legal move is executed, the player loses control, which we specify in our framework as:
\begin{lstlisting}
abnormal(control(P), choice(P, M), S).
\end{lstlisting}
I.e., after a choice made by a player, that player loses control and cannot play a move again from that situation onwards. Moves made in this way bring about the final situation, which we specify as a situation term with two choices made from the initial situation, one for each player.
\begin{lstlisting}
final(S):-
    ground(S),
    S=do(choice(_,_), do(choice(_,_), I)),
    initial(I).
\end{lstlisting}
Assuming the payoff matrix of Table~\ref{subtab:PD} defined as:
\begin{lstlisting}
payoff('D', 'D', 35, 35).
payoff('C', 'D', 10, 100).
payoff('D', 'C', 100, 10).
payoff('C', 'C', 65, 65).
\end{lstlisting}
the outcome of the game holds information about the actual moves made by the players and their payoffs.
\begin{lstlisting}
finally(outcome(P1,M1,U1,P2,M2,U2), S):-
    final(S),
    holds(role(P1, row), S),
    holds(did(P1, M1), S),
    holds(role(P2, col), S),
    holds(did(P2, M2), S),
    payoff(M1, M2, U1, U2).
\end{lstlisting}
We can extract outcome information, e.g. the utility for a player, through a {\tt goal/2} fluent (as in GDL) e.g.:
\begin{lstlisting}
finally(goal(P1, U1), S):-
    finally(outcome(P1,_,U1,_,_,_), S).
finally(goal(P2, U2), S):-
    finally(outcome(_,_,_,P2,_,U2), S).
\end{lstlisting}
This completes the definition of a PD game and allows us to use the above game description to reason about the game. For instance, if player {\tt p1} wanted to reason about the game and determine the actions required to achieve a utility of 100, expressed as the query:
\begin{lstlisting}
?- game(s0, F), finally(goal(p1, 100), F).
\end{lstlisting}
based on the game's payoff matrix, our solver provides the following two solutions:
\begin{lstlisting}
F=do(choice(p2,'C'),do(choice(p1,'D'),s0)) ;
F=do(choice(p1,'D'),do(choice(p2,'C'),s0)) ;
false.
\end{lstlisting}
In the first answer for {\tt F} where {\tt p1} acted first and {\tt p2} second, while in the second answer, the order is reversed. This is not unexpected, as the way we have defined the game both players have control in the initial state, and the answers show both combinations.

\subsection{Framework}

\begin{algorithm}
	\small
	\caption{Generating game-specific predicates from natural language descriptions. PD stands for Prisoner's Dilemma.}
	\begin{algorithmic}[1]
		\State \textbf{Input:} \newline \texttt{$\Gamma$}: game-independent predicates, \newline \texttt{$NL_{PD}$}: natural language description of PD, \newline \texttt{$\xi_{PD}$}: game-specific predicates of PD, \newline \texttt{$NL_{NG}$}: natural language description of a new game. 
		\State \textbf{Output:} \newline \texttt{$\xi_{NG}$}: game-specific predicates for the new game. 
		\State \textbf{Parameter:} \newline \texttt{$\text{max\_attempts}$}: maximum correction attempts.
		
		\State \texttt{$\text{attempts} \gets 0$}
		\State \texttt{$\text{trace} \gets \emptyset$}
		
		\While{\texttt{$\text{attempts} < \text{max\_attempts}$}}
		\State \texttt{$\xi_{NG} \gets \text{LLM.translate}(\Gamma, NL_{PD}, \xi_{PD}, NL_{NG})$}
		\State \texttt{$\text{is\_valid} \gets \text{solver.check\_predicates}(\xi_{NG})$}
		
		\If{\texttt{$\text{is\_valid}$}}
		\State \Return \texttt{$\xi_{NG}$}
		\Else
		\State \texttt{$\text{trace} \gets \text{solver.get\_trace}()$}
		\State \texttt{$\xi_{NG} \gets \text{LLM.self\_correct}(\xi_{NG}, \text{trace})$}
		\EndIf
		\State \texttt{$\text{attempts} \gets \text{attempts} + 1$}
		\EndWhile
		
		\State \Return \texttt{Unable to generate valid predicates within maximum attempts.}
		
	\end{algorithmic}
	\label{alg:algorithm}
\end{algorithm}

To generate a formal game specification for a given game, we use the one-shot prompting approach. The prompt consists of game-independent predicates (\texttt{$\Gamma$}), an example: a natural language description of Prisoner's Dilemma (\texttt{$NL_{PD}$}) and game-specific predicates for PD (\texttt{$\xi_{PD}$}), and a natural language description of a game to be translated (\texttt{$NL_{NG}$}). An LLM is prompted iteratively. Once the game-specific predicates (\texttt{$\xi_{NG}$}) are generated, they are validated for syntactic correctness using a Prolog solver. If the generated predicates contain syntax errors, the LLM is re-prompted with the solver's error trace and additional instructions for correcting the Prolog code. This correction process is repeated up to \texttt{max\_attempts}. An overview of the algorithm is provided in Listing~\ref{alg:algorithm}.

\subsection{Game descriptions}

\begin{table}[htb]
	\centering
	\begin{tabular}{c|cc}
		\textbf{Payoffs/Description} & \textbf{Standard} & \textbf{Non-standard} \\ \hline
		\textbf{Numerical} & 5 & 50 \\ 
		\textbf{Non-numerical} & 5 & 50 \\
	\end{tabular}
	\caption{Number of natural language game descriptions for each variant. \textbf{Standard} refers to example employing a typical metaphor for a given game. Conversely, \textbf{non-standard} is a newly invented example of a situation that can be modelled by a given game. \textbf{Numerical} refers to descriptions containing numerical payoffs, and \textbf{non-numerical} to descriptions without numerical payoffs.}
	\label{tab:variants}
\end{table}

To evaluate the ability of \gpt{} to translate natural language game descriptions into their formal specification, we used standard examples of five classic 2-player simultaneous one-shot games from game theory: Battle of the Sexes, Hawk-Dove, Matching Pennies, Prisoner's Dilemma, and Stag Hunt. Additionally, we investigated the \gpt{}'s capability to formalize a description of a real-world conflict or cooperation situation that can be modelled by one of the five games but differs from the typically used metaphor.

The real-world dilemma descriptions were synthetically generated. For each game, a real-life scenario was developed that avoided the standard metaphor (e.g., avoiding two prisoners for the Prisoner’s Dilemma). Then, \gpt{} was prompted to create further dilemmas from various domains such as social life, business, politics, and warfare, using the initial scenario as a template. To increase the difficulty, some natural language descriptions were generated without numerical payoffs. Table~\ref{tab:variants} summarizes the number of game descriptions for each variant, resulting in 110 descriptions of simultaneous move games.        

In a separate experiment, we evaluated the performance of \gpt{} using a zero-shot prompting approach. In this case, only the game-independent predicates (\texttt{$\Gamma$}) were provided in the prompt, without the inclusion of any example game-specific predicates, such as \texttt{$\xi_{PD}$}. This experiment aimed to assess the model’s ability to formalize a game without prior examples. Additionally, to analyse the LLM’s capacity to generalize from examples, we conducted an autoformalization of two differently structured games: a sequential version of the Prisoner’s Dilemma, where players take turns making decisions, and the Rock-Paper-Scissors game, which provides more than two choices per player and lacks specified numerical payoffs.

\subsection{Experimental parameters}

\begin{table}[h!]
	\centering
	\begin{tabular}{|l|l|}
		\hline
		\texttt{LLM} & \gpt{} \\ \hline
		\texttt{temperature} & $1$ \\ \hline                                      
		\texttt{maximum output tokens} & $1024$ \\ \hline
		\texttt{maximum attempts number} & $5$ \\ \hline
	\end{tabular}
	\caption{Experimental parameters. }
	\label{tab:exp-parameters}
\end{table}

To perform autoformalization we employed state-of-the-art model \gpt{}. The experimental parameters used in the evaluation are shown in Table~\ref{tab:exp-parameters}. The source code and the logs generated in the experiment are publicly available\footnote{\color{blue}{\url{https://github.com/dicelab-rhul/game-formaliser}}}. 

\section{Results and Discussion}
\label{sec:results}

Our experimental results include both zero-shot and one-shot prompting, as well as testing the model's capability to generate formal specifications for games that differ structurally from the provided examples. Although the outcomes from zero-shot prompting demonstrate the weaknesses of LLMs to produce formal description for our game-independent part, one-shot prompting with an example is able to provide results that adapt successfully the two-player game formulations for different game instances.

\subsection{Zero-shot prompting}

We started by performing zero-shot prompting, including only the predicates for the game-independent part in the prompt. The goal was to generate the game-dependent part from a textual description without an example. For PD, the following code sequence was generated as an answer. The first part of the answer contained the representation of the players and the legal moves:

\begin{lstlisting}
suspect(suspect1).
suspect(suspect2).

legal(confess(S), _):- suspect(S).
legal(silent(S), _):- suspect(S).
\end{lstlisting}
Although players represented as suspects is acceptable, players' moves being legal in any state is not; the underscore ({\tt '\_'}) indicates that state is unnecessary when checking whether a move is legal. This is obviously undesired, as it is illegal to make two consecutive moves, but the formulation allows it. The generated answer also contains what holds initially:

\begin{lstlisting}
initially(state(0, 0), s0).
\end{lstlisting}
Here clearly the generated code is problematic as the term {\tt state(0,0)} has unclear parameters, including their {\tt 0} value. Also, problematic is the code generated subsequently, for when a state is final:
\begin{lstlisting}
final(state(_, _)).
\end{lstlisting}
This implies that any state term of the form {\tt state(\_,\_)} is final. So already the solution so far is unusable. The situation deteriorates with the effects of moves. We show below an effect rule, specified as:

\begin{lstlisting}
effect(state(Y1, Y2), silent(suspect1), state(NY1, NY2)) :- 
    holds(state(Y1, Y2), s0), 
    ( % If suspect2 also silent
        holds(state(_, _), do(silent(suspect2), s0)) 
    -> NY1 is Y1 + 1, NY2 is Y2 + 1
    ; % If suspect2 confesses
      NY1 is Y1, NY2 is 10
    ).
\end{lstlisting}
The above effect rule has the state as the first and third arguments, but the state is not in situation calculus format. As a result, only what happens at the initial state will be inferred correctly. Also, the first argument is intended as a fluent in the game-independent part of our formulation, but the generated answer does not reflect this. Moreover, from the above, we can now see that the two parameters of the state are used for storing the players' utility, originally set to zero, updated wrongly, and missing all other useful state information that is required. Finally, this kind of formulation for effects requires that nothing can be abnormal in the generated answer:
\begin{lstlisting}
abnormal(_, _, _) :- false.
\end{lstlisting}
All the above show that there is no deep understanding of how the situation calculus really works.

\subsection{2-player simultaneous move games}

\begin{table}
	\centering
	\begin{tabular}{c|cccc}
		\textbf{Payoffs/Desc.} & \multicolumn{2}{c}{\textbf{Standard}} & \multicolumn{2}{c}{\textbf{Non-standard}} \\
		& \texttt{synt} & \texttt{sem} & \texttt{synt} & \texttt{sem} \\ \hline
		\textbf{Numerical} & 1.00 & 1.00 & 1.00 & 0.90  \\ 
		\textbf{Non-numerical} & 1.00 & 1.00 & 0.96 & 0.83 \\
	\end{tabular}
	\caption{Syntactic and semantic accuracy of generated game-specific predicates. \texttt{Synt} denotes syntactic accuracy, \texttt{sem} denotes semantic accuracy.}
	\label{tab:accuracy}
\end{table}

In the one-shot prompting method, we provided game-specific axioms for prisoner's dilemma, $\xi_{PD}$, as an example. This resulted in the generation of game-specific predicates for the natural language description variants presented in Table~\ref{tab:variants}, analogous to those defined for PD (see Sect.~\ref{subsec:solver}). The main difference was the definition of the payoff matrix. Table~\ref{tab:accuracy} summarizes the syntactic and semantic accuracy of the generated predicates. Syntactic accuracy refers to the percentage of syntactically correct Prolog programs produced on the first translation attempt, before any revisions were made. The semantic correctness of the generated predicates (e.g. the relationships between payoffs in the payoff matrix) was reviewed manually. During the inspection of non-standard natural language descriptions, some instances were found to be incomplete or ambiguous. As a result, semantic accuracy was calculated based on 48 samples for the numerical variants and 47 samples for the non-numerical variants.      

The generated code was syntactically correct in 108 out of 110 cases (98\%). In one of the incorrect cases, the error involved the use of a comment delimiter ('//'), not valid in Prolog. In the second case, the code triggered a warning about singleton variables. Notably, despite receiving feedback from the Prolog solver, \gpt{} was unable to fix the syntactic errors. This contrasts with our previous work, where the model was able to correct reasoning errors in game-theoretic scenarios based on solver feedback~\cite{mensfelt2024lelma}. A potential explanation could be that the feedback prompt, which included the solver trace and debugging guidelines, may not have been specific enough. However, when the model was re-prompted in a separate call, syntactically correct code was generated on the first attempt. For analyzing semantic correctness, the syntactically incorrect codes were replaced with corrected versions.

\begin{figure}
	\centering
	\includegraphics[width=\columnwidth]{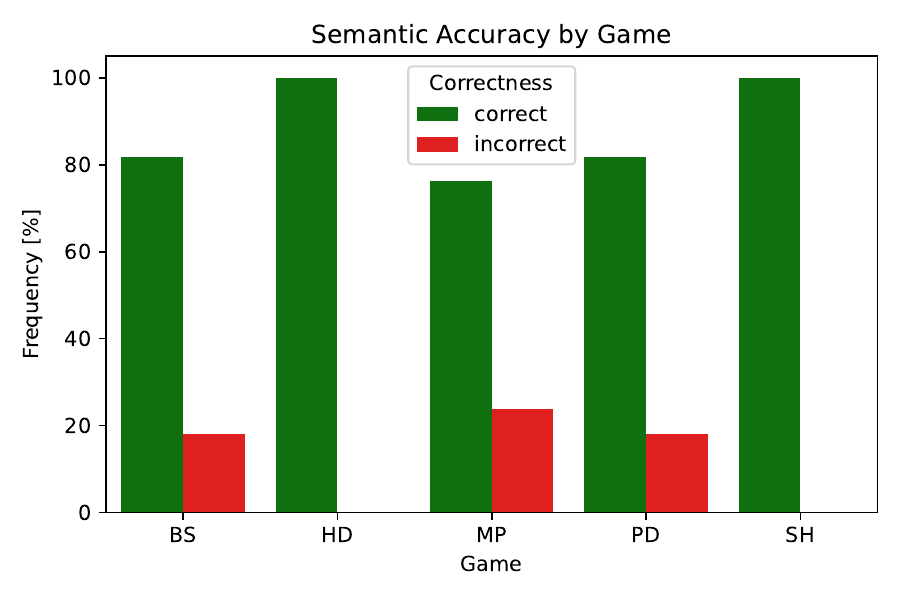}
	\caption{Semantic accuracy of generated predicates for each of the games. ``BS'', ``HD'', ``MP'', ``PD'', and ``SH'' stand for Battle of the Sexes, Hawk-Dove, Matching Pennies, Prisoner's Dilemma, and Stag-Hunt, respectively.}
	\label{fig:games-accuracy}
\end{figure}

Regarding semantic correctness, the standard examples were translated into their formal specification with 100\% accuracy. For the non-standard examples, the accuracy was 90\% for the numerical variants and 83\% for the non-numerical ones. Fig.~\ref{fig:games-accuracy} shows the semantic accuracy for each of the considered games. Interestingly, even though the Prisoner's Dilemma was provided as an example in the prompt, some translations of real-world situations modeled by PD were incorrect. In contrast, all translations for Hawk-Dove and Stag Hunt were semantically accurate, despite no specific examples being provided in the prompt for these games. It may suggest that the Prisoner's Dilemma is more challenging, however, the number of errors was not significantly high. Most of the semantic errors involved incorrect payoff assignment in the predicates specifying the payoff matrix. In one case, one, instead of two constants was used in predicates describing the payoff matrix:

\vspace{0.5em}

\noindent \texttt{payoffRD('Holiday\_Holiday', -1, 3)}

\vspace{0.5em}

Overall, the translation showed high syntactic and lower semantic correctness. It is worth noting that the real-world examples were translated in most cases correctly, even in the absence of numerical payoffs.

\subsection{Towards Generalisation}

{\bf Sequential PD} In order to explore whether the LLM can generalise over different aspects of a game, we also run our system to see what is generated in the case of the sequential PD. This is a turn-based two-player game in which there is a first-mover that chooses to cooperate or defect, and after observing this choice, the second-mover responds with an action of the same set. In terms of the game specification, the LLM should produce the same specification as the game-dependent part of Section~\ref{subsec:solver}, with the only differences (a) a revised payoff matrix and (b) the way the initial state is specified, to allow only for one player to play first, and not both, i.e. it should contain only one player having the control of the game at state {\tt s0}. Indeed, as shown below, the system generated in the initial state two players {\tt a} and {\tt b}, the role of {\tt a} is to play first, and of {\tt b} second, and the control is given to only to {\tt a}:

\begin{lstlisting}
initially(player(a), s0).
initially(player(b), s0).
initially(role(a, first), s0).
initially(role(b, second), s0).
initially(control(a), s0).
\end{lstlisting}
{\bf Rock-Paper-Scissors} We also explored whether the LLM can generalise over the number of moves of the game and the specification of the possible moves. For this purpose, we gave it the textual description of the Rock-Paper-Scissors game, a simultaneous zero-sum game that has three possible outcomes: a draw, a win, or a loss. A player who decides to play rock will beat another player who chooses scissors (``rock crushes scissors''), but will lose to one who has played paper (``paper covers rock''); a play of paper will lose to a play of scissors (``scissors cuts paper''). If both players choose the same shape, the game is tied and is usually replayed to break the tie. Again, our system produced the same specification as the game-dependent part of Section~\ref{subsec:solver}, with the only differences (a) a revised payoff matrix and (b) the way possible moves were specified, to allow for three moves instead of two, as shown below:

\begin{lstlisting}
possible(choice(P, 'rock'), S) :- 
    holds(player(P), S).
possible(choice(P, 'paper'), S) :- 
    holds(player(P), S).
possible(choice(P, 'scissors'), S) :- 
    holds(player(P), S).
\end{lstlisting}
This demonstrates that our original formulation of a domain-specific game allowed our system to generalise well to different aspects of a new game.

\section{Related Work}
\label{sec:related-work}

LLMs have gained significant interest as potential building blocks for agents in game-theoretical simulations~\cite{akata2023playing,fan2023can,guo2023gpt,lore2023strategic}. Despite their promise, a major obstacle is the inherent challenge that reasoning tasks pose for LLMs~\cite{rae2022scaling}, \cite{nye2021improving}. While techniques such as Chain-of-Thought~\cite{wei2023chainofthought} can enhance performance in some of the tasks, even state-of-the-art models are not free from logical and arithmetic errors~\cite{imani2023mathprompter}. An alternative approach to leveraging LLMs for reasoning in game-theoretic scenarios is to utilise them for translating natural language into formal representations, followed by employing a formal solver for the actual reasoning process. This method, known as autoformalization~\cite{wu2022autoformalization}, has been successful in translating natural language into mathematical formalisms~\cite{wu2022autoformalization,he2023solving,jiang2022draft}, providing a method to integrate LLMs with formal tools to solve reasoning tasks.

The autoformalization approach was also successfully applied to creating logical representations of natural language descriptions. Yang et al.~\cite{yang2023coupling} used \texttt{GPT-3} and few-shot prompting to translate natural language sentences into their formal representation, passed as an input to answer set programs. \texttt{GPT-3} autoformalization coupled with ASP achieved high accuracy on NLP benchmarks, however, errors in translation from natural language occurred in some cases. Pan et al.~\cite{pan2023logic} combined in-context learning with self-refinement based on the evaluation from the solver. This improved the performance over standard LLM and chain-of-thought prompting on logical reasoning benchmarks. This approach was also not free from translation errors. To address this problem in the context of translation to temporal logics, Cosler et al.~\cite{cosler2023nl2spec} introduced an interactive approach in which the user can edit sub-translations. Fine-tuning of LLMs was demonstrated to be another successful method for increasing translation accuracy in both temporal logics~\cite{chen2023nl2tl} and deductive datasets~\cite{feng2023language}.

The aforementioned works considered standard logical reasoning tasks and benchmarks. To the best of our knowledge, this work is the first to perform autoformalization in the context of game theory. 

\section{Conclusions and Future Work}
\label{sec:conclusions}

To evaluate the ability of our framework and \gpt{} to perform autoformalization in the domain of logical representation of game-theoretic scenarios, we developed a dataset of natural-language descriptions. The descriptions involved scenarios ranging from family decision-making to business to warfare, possible to model by 2-players simultaneous-move games. The descriptions were developed in two versions, including and lacking numerical payoffs.

The preliminary assessment demonstrated that with zero-shot prompting \gpt{} was not able to autoformalize the standard description of the Prisoner's Dilemma. One-shot prompting, however, showed high syntactic and semantic accuracy. Interestingly, even in the case of real-world examples with no numeric payoffs, the LLM was mostly able to infer an appropriate payoff matrix. To evaluate the ability of generalization within the proposed framework, we used the natural-language descriptions of sequential Prisoner's Dilemma and Rock-Paper-Scissors. The formal representation of both games, differing in structure from non-sequential PD provided as an example in the prompt, were syntactically and semantically correct.

We acknowledge several limitations of this work. The semantic correctness was investigated manually, which limits the scalability of the evaluation. We plan to address this in future work by not only generating the predicates for the game description but also executing the code to find payoffs for specific actions, which will allow us to automatize the evaluation. The class of games that we considered was limited to 2-players simultaneous-move games and two games that differ only in the sequential nature of decision-making and the number of possible actions. We will expand the set of considered games to include games from outside this class and outside game theory.    

Overall, the results show the promise of using LLMs as a bridge between real-life strategic interactions and the game theory framework, facilitating the use of formal tools in finding an optimal action in such scenarios.

\section*{Acknowledgments}

This work was supported by a Leverhulme Trust International Professorship Grant (LIP-2022-001).

\bibliographystyle{kr}
\bibliography{references}

\end{document}